\documentclass{article}

\usepackage[final, nonatbib]{nips_2017}

\usepackage[utf8]{inputenc} 
\usepackage[T1]{fontenc}    
\usepackage{url}            
\usepackage{booktabs}       
\usepackage{amsmath, amsfonts, gensymb}
\usepackage{nicefrac}       
\usepackage{microtype}      

\usepackage{graphicx}			
\usepackage{subfig}

\title{Data Augmentation for Detection of Architectural Distortion in Digital Mammography using Deep Learning Approach}

\author{
	Arthur C. Costa$^1$\textnormal{, }Helder C. R. Oliveira$^1$ \\
    \textbf{Juliana~H.~Catani}$^2$, \textbf{Nestor~de~Barros}$^2$, \textbf{Carlos~F.~E.~Melo}$^3$\\
    \textbf{Marcelo A. C. Vieira}$^1$\thanks{\emph{Corresponding author}: \texttt{mvieira@sc.usp.br}.} \\
    $^1$Dept. of Electrical Engineering, University of São Paulo, São Carlos, Brazil \\
    $^2$Dept. of Radiology, University of São Paulo, São Paulo, Brazil\\
    $^3$Eco \& Mama Diagnóstico Digital, São Carlos, Brazil \\
   }

\begin{document}

\maketitle

\begin{abstract}
Early detection of breast cancer can increase treatment efficiency. Architectural Distortion (AD) is a very subtle contraction of the breast tissue and may represent the earliest sign of cancer. Since it is very likely to be unnoticed by radiologists, several approaches have been proposed over the years but none using deep learning techniques. To train a Convolutional Neural Network (CNN), which is a deep neural architecture, is necessary a huge amount of data. To overcome this problem, this paper proposes a data augmentation approach applied to clinical image dataset to properly train a CNN. Results using receiver operating characteristic analysis showed that with a very limited dataset we could train a CNN to detect AD in digital mammography with area under the curve (AUC = 0.74).
\end{abstract}

\section{Introduction}
Breast cancer is the most lethal cancer among women worldwide~\cite{oms}. Treatment efficiency can increase up to 30\% if detected in earlier stages ~\cite{Veronesi2005}. To detect those tumors the most adopted exam is the Full-field Digital Mammography (FFDM), where the radiologist looks for anomalies such as masses, microcalcifications and architectural distortion~\cite{Glynn2011effect}. Architectural Distortion (AD) is the earliest manifestation of breast cancer and can appear up to two years before the formation of any other anomaly. It is a very subtle anomaly that changes the texture of breast parenchyma and it is very hard to be detected through the human visual system~\cite{Bahl2015}. It has been reported that AD is the most common finding in retrospective cases of false-negative in FFDM ~\cite{Bahl2015,Rangayyan2010}.

Deep Learning (DL) is a relatively new area of Machine Learning (ML) that have attracted the attention of the scientific community. Its great advantage of methods simplify the whole classical process of description and pattern classification, by replacing it with a general learning procedure that provides information at various levels of abstraction~\cite{lecun2015deep}. As a result, some Computer Aided Diagnosis (CAD) systems are currently using deep architectures to improve diagnostic accuracy, which has yielded superior results if compared to those using classical ML systems~\cite{greenspan2016guest}. 

One of the most popular deep architectures is the Convolutional Neural Network (CNN)~\cite{lecun2015deep}. Its arrangement is basically structured of overlapping convolution and pooling layers that are responsible for the pattern learn which occurs at various levels of abstraction. The training process of this network has become increasingly practical thanks to its parallelizable algorithm allied to technological advances in graphical processing units, making it the most popular deep architecture currently used for developments with CAD~\cite{greenspan2016guest}.

One implication of using deep architectures is the high amount of data it requires to perform a proper model training. To overcome this limitation, a technique called \emph{Data Augmentation} has been widely used to increase the dataset samples, by applying transformations on the input samples to increase the amount of data~\cite{greenspan2016guest}. 

This paper focus on the use of DL to help the detection of AD in FFDM images. Because clinical FFDM datasets are most private and, when available, with limited amount of cases with AD, this paper proposes the use of data augmentation to improve the training step of the CNN even with a limited number of images in the dataset.


%

\section{Material and Methods}

\subsection{Mammography Dataset}
In this work the initial dataset is composed of 300 clinical images, where 200 are digitized mammograms from the freely available DDSM dataset~\cite{DDSM} and 100 are digital mammography (DM) images obtained under review board approval\footnote[1]{CAAE $\#56699016.7.0000.0065$}. The equipment used to acquire the DM images was a Hologic Selenia Dimensions\textsuperscript{\textregistered} mammography system. All images have AD marked by an experienced radiologist.


The selected images were cropped based on the radiologist marks. In order to have a sample of normal tissue, another random region was selected in the same image, resulting in a total of 600 squared Regions of Interest (ROIs) with same size after a preprocessing step. 




\subsection{Data Augmentation}
Since a small amount of samples is a constraint for training deep CNNs, data augmentation was performed over the initial ROIs aiming the dataset expansion. To do so, each cropped ROI passed through some transformations: vertical and/or horizontal flip; rotation at 90, 180, and 270 degrees; addition of Gaussian noise with zero mean and 0.02, 0.04 and 0.06 variance. Some transformations were a composition of one or more steps pointed before and, by the end of the augmentation, the new dataset had 21600 samples of balanced classes.

\subsection{Convolutional Neural Network}
The CNN algorithm was implemented using the TensorFlow\textsuperscript{\textregistered} framework created by Google\textsuperscript{\textregistered}. The network architecture was designed so that the final convolutional layer provided a down sampled $4x4$ feature map to be delivered to the fully-connected layer. A set of $5x5$ filters were trained in each layer with quantities scaling by the factor of two. For pooling purposes, the max pooling function was used with no overlaps over the features maps.

To perform the training of the CNN, the dataset was split into 70\%-15\%-15\% (training-validation-test) with equally divided classes. The input images were standardized by \emph{zscore} algorithm. For each training step, a 60 samples size batch was fed into the input layer.

\section{Results}
\begin{figure}[h]
	\centering
	\subfloat[]{
        \includegraphics[height=120px]{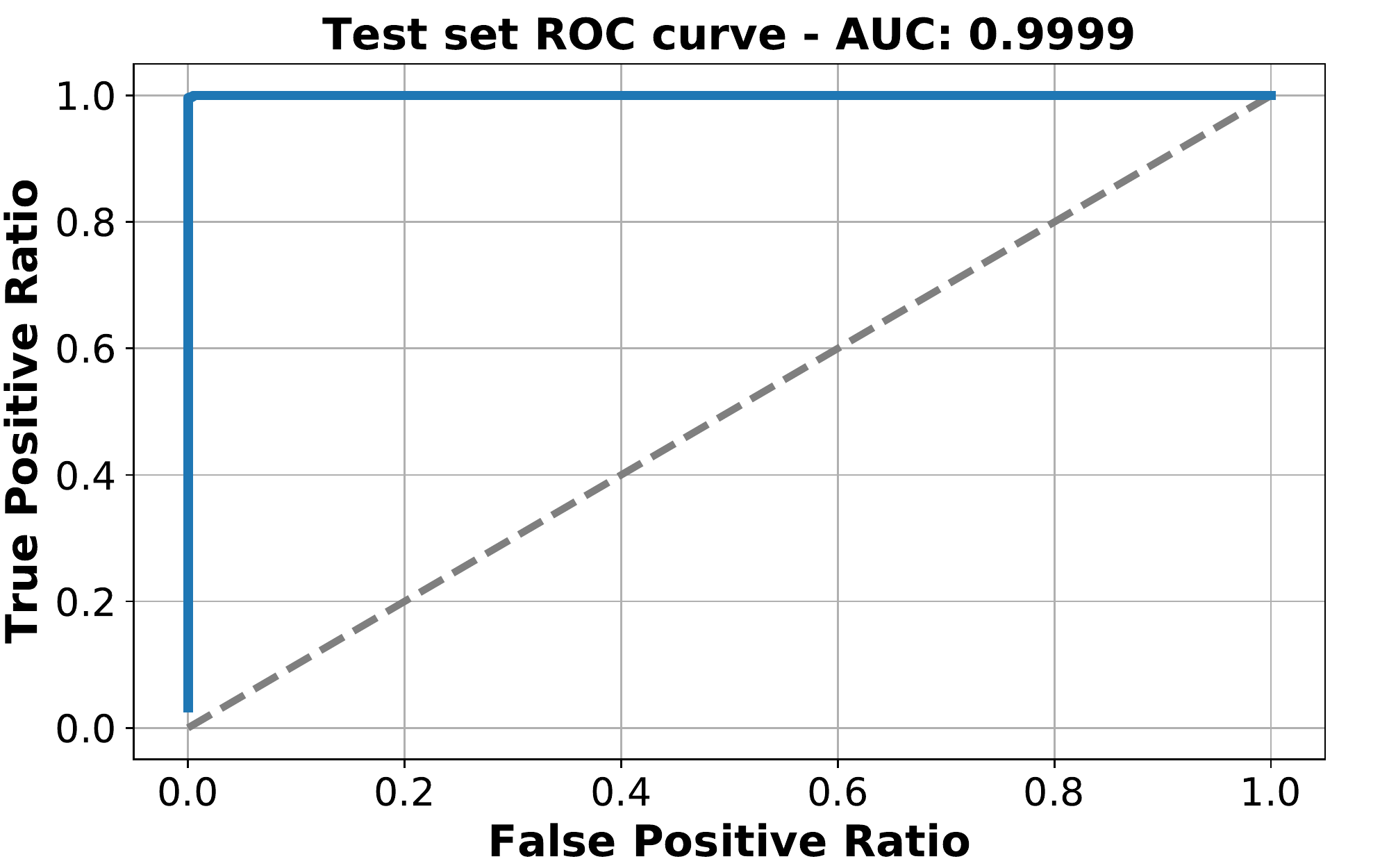}
        \label{ROC_Test}
    }    
    \subfloat[]{
        \includegraphics[height=120px]{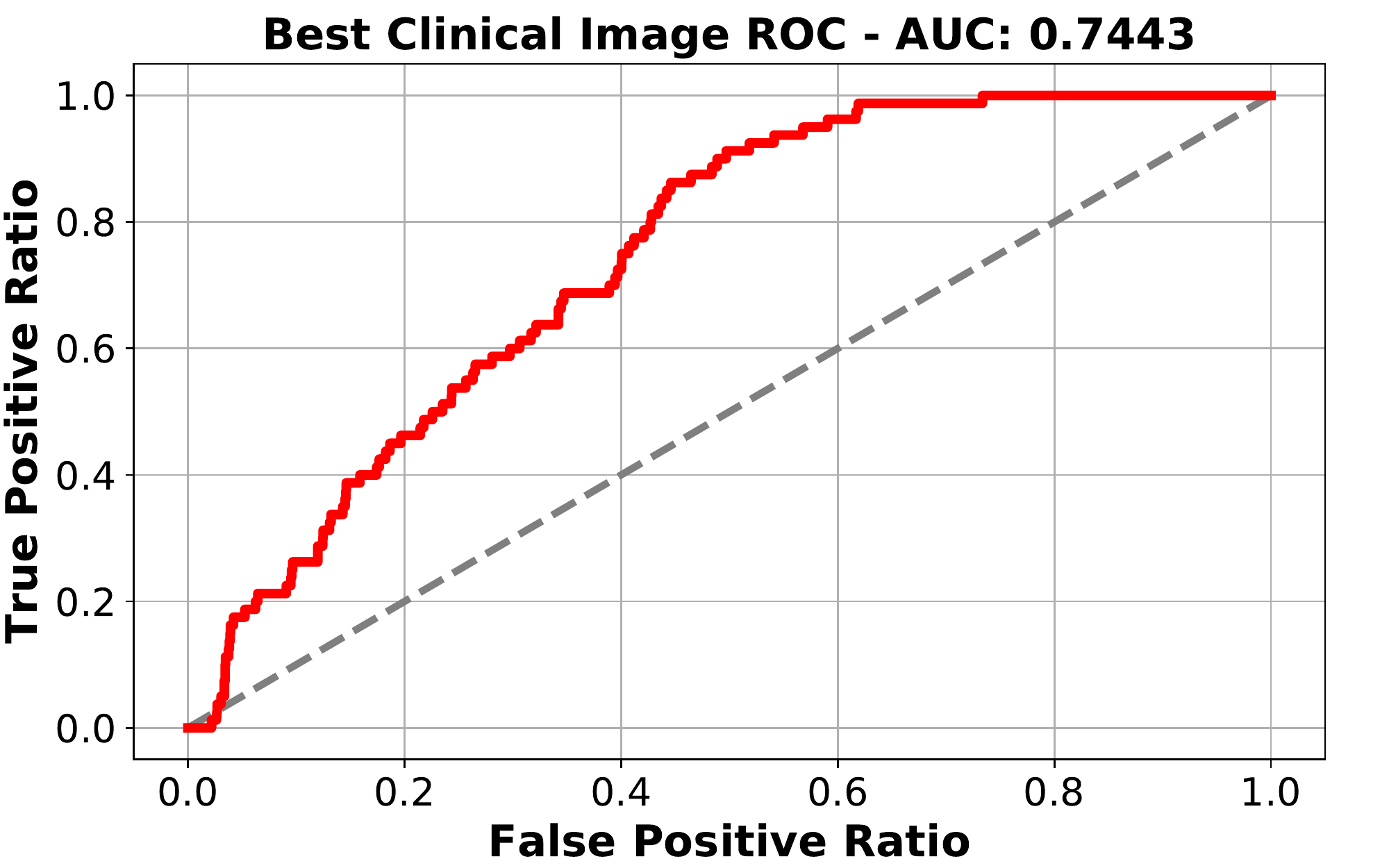}
        \label{ROC_BestClinical}
    }  
	\caption{ROC curve of the (a) model training  and (b) test using a realistic approach.}
	\label{fig:roc}
\end{figure}

The CNN training process took into account the cost function variation, and by the end of the training, the model was applied on the test set obtaining 99,4\% of accuracy. The  respective Receiving Operator Characteristic (ROC) curve is visualized at Figure~\ref{ROC_Test}, where the Area Under the Curve (AUC = 0.99). This result shows that the CNN almost got a perfect performance on the test samples.

To evaluate the trained model in a more realistic clinical scenario, nine new exams were used for validation. Several ROIs were extracted from the segmented breast area in the images. The extraction was performed by attempting to cover all the possible regions where several $256x256$ ROI could be gathered from. Each exam provided about 3000 ROIs, which were labeled as AD if any pixel of the ROI contained the AD center coordinates, and labeled as normal tissue otherwise. The trained CNN model was then fed with the new data, and the exam whose CNN results a higher AUC is presented at Figure~\ref{ROC_BestClinical}, with accuracy of 86,1\%.

\section{Conclusions and Future Works}
This paper has proposed a data augmentation technique to detect architectural distortion in breast images using deep learning. Such a lesion is very important for early diagnosis of breast cancer. The results reach the best case of an AUC=0.74. Since the main limitation of the proposed approach is to manually crop the ROIs used to data augmentation and consequently to train the CNN, our efforts will focus on the automation of this step. The dataset size is a limitation as well, which we intend to expand in the near future.

\section*{Acknowledgments}
This project is supported by São Paulo Research Foundation (FAPESP) grant $\#2015/20812-5$, the National Council for Scientific and Technological Development (CNPq), the Coordination for the Improvement of Higher Education Personnel (CAPES). We also acknowledge the support of NVIDIA Corporation with the donation of the Quadro M5000 GPU used for this research.

\bibliographystyle{unsrt}
\bibliography{bibliografia}
\end{document}